\documentclass[conference]{IEEEtran}
\IEEEoverridecommandlockouts
\usepackage{float}
\usepackage{cite}
\usepackage{amsmath,amssymb,amsfonts}
\usepackage{algorithmic}
\usepackage{graphicx}
\usepackage{textcomp}
\usepackage{xcolor}
\usepackage{hyperref}
\usepackage{mathtools}
\usepackage{amsmath}
\def\BibTeX{{\rm B\kern-.05em{\sc i\kern-.025em b}\kern-.08em
    T\kern-.1667em\lower.7ex\hbox{E}\kern-.125emX}}
\usepackage[autostyle]{csquotes}

\begin{document}

\title{Robot Imitation from Video Demonstration

\thanks{This project is submitted as part of ASU's CSE 598: Advances in Robot Learning}
}

\author{\IEEEauthorblockN{ Venkat Surya Teja Chereddy}
\IEEEauthorblockA{
\textit{Arizona State University}\\
Tempe, USA \\
vcheredd@asu.edu}
}

\maketitle

\begin{abstract}
This paper presents an attempt to replicate the robot imitation work conducted by Sermanet et al. \cite{tcn}, with a specific focus on the experiments involving robot joint position prediction. While the original study utilized human poses to predict robot joint positions, this project aimed to achieve robot-to-robot imitation due to the challenges of obtaining human-to-robot translation data. The primary objective was to provide a neural network with robot images and have it predict end-effector positions through regression. The paper discusses the implementation process, including data collection using the open-source RoboSuite, where a Python module was developed to capture randomized action data for four different robots. Challenges in data collection, such as oscillations and limited action variety, were addressed through domain randomization. Results show high testing error and unsatisfactory imitation due to overfitting, necessitating improvements in the project.
\end{abstract}

\section{What I Set Out to Do}
The goal in my mind, when I first started this project, is to replicate the work of Sermanet et al.\cite{tcn},
specifically the robot imitation from the experiments section. In the original paper, the authors have used human poses to predict the joint positions of the robot. I, 
however, did not plan for a human-to-robot translation for this project. This is primarily because I need
to manually collect the human poses' to joint spaces data, which is hard given that this is a solo project.
Therefore, I planned for robot-to-robot imitation, since the collection of data can be simpler for simulation data. In simple terms, the idea of this project is to give a neural network the image of the robot, and the network predicts the end-effector positions, using regression. 

For the training process, I first learn a TCN representation space, using the triplet loss implementation. Once the TCN representation space is learned, I train a feed forward
regression network that maps the TCN embeddings to the end-effector positions 
(it should be noted that the original paper is trained on the joint-space instead of end-effector positions) of the robot. 

\section{Implementation (and the Difficulties I Faced)}

\subsection{Collecting Data: Part 1}

For this project, I need quality data of robots (preferably more than 3 robots) performing random actions.
After researching the internet, I found an open-source project: RoboSuite. This project is primarily used for Reinforcement Learning research and gives a Gym like API. The project also implements random action, 
however, the random actions are not good as they produce oscillations and the actions are limited.

I built a Python Module with RoboSuite in the backend, for the random data collection. This Python Module
supports domain randomized, random action data collection for 4 robots: Panda, Sawyer, IIWA and Jaco. 
This Python Module saves the image and end-effector positions at every timestep. On top this, this Python
module supports action replay, given an array of end-effector positions. 

\subsection{Training TCN and Regression: Part 1}
During the first time I trained, I made two big mistakes:
\subsubsection{Not Using a Validation Dataset}
Not using a validation dataset has made it very hard for me to know if the TCN is overfit, which is surely did to some extent, when I checked the TCN representation space on a new dataset.

\subsubsection{Regression Never Converged Because I Collected the Wrong Data}
Unknown to me, the RoboSuite library uses delta changes as inputs and outputs for its end-effector poses for action/replay. When I tried to train using this data, the network never converged, as expected.
All the data I have collected is, therefore, unfortunately, useless.

\subsection{Collecting Data: Part 2}
I made changes to my Python module for it to work with global positions, instead of delta positions at each time step. I have collected 80 videos (120 frames each) of the four robots with domain randomization at every 60 frames.

\subsection{learning TCN and Regression: Part 2} \label{workingtcn}
This time I have split the main dataset into training and validation datasets. Instead of first training
TCN and then training regression, I have used a different approach: training them together. After trail and
error I have used the following loss function:

\equation{\mathcal{L}_{TCN}= max(Dist(\phi(A),\phi(P)) - Dist(\phi(A),\phi(N)) + m, 0)}
$$

\equation{\mathcal{L}_{Regression}= MSE(J_A, j_A) + MSE(J_p, j_p) + MSE(J_n, j_n)}
$$
\equation{\mathcal{L}_{total} = 0.4 * \mathcal{L}_{TCN} + \mathcal{L}_{Regression}}
$$
\textit{Where, $\phi(x)$ = TCN network, A = Anchor image, P = Positive image, N = Negative image, m = margin (Hyperparameter, I set it to 5),  Dist = Euclidean distance,  J = Predicted joint position, j = actual joint position}

The network has started converging after making these changes, however, it soon began to over-fit the training data, so I had to stop early. I believe this is because of relatively small data-set size.

\section{Results}
\begin{figure}[H]
\includegraphics[scale=0.15]{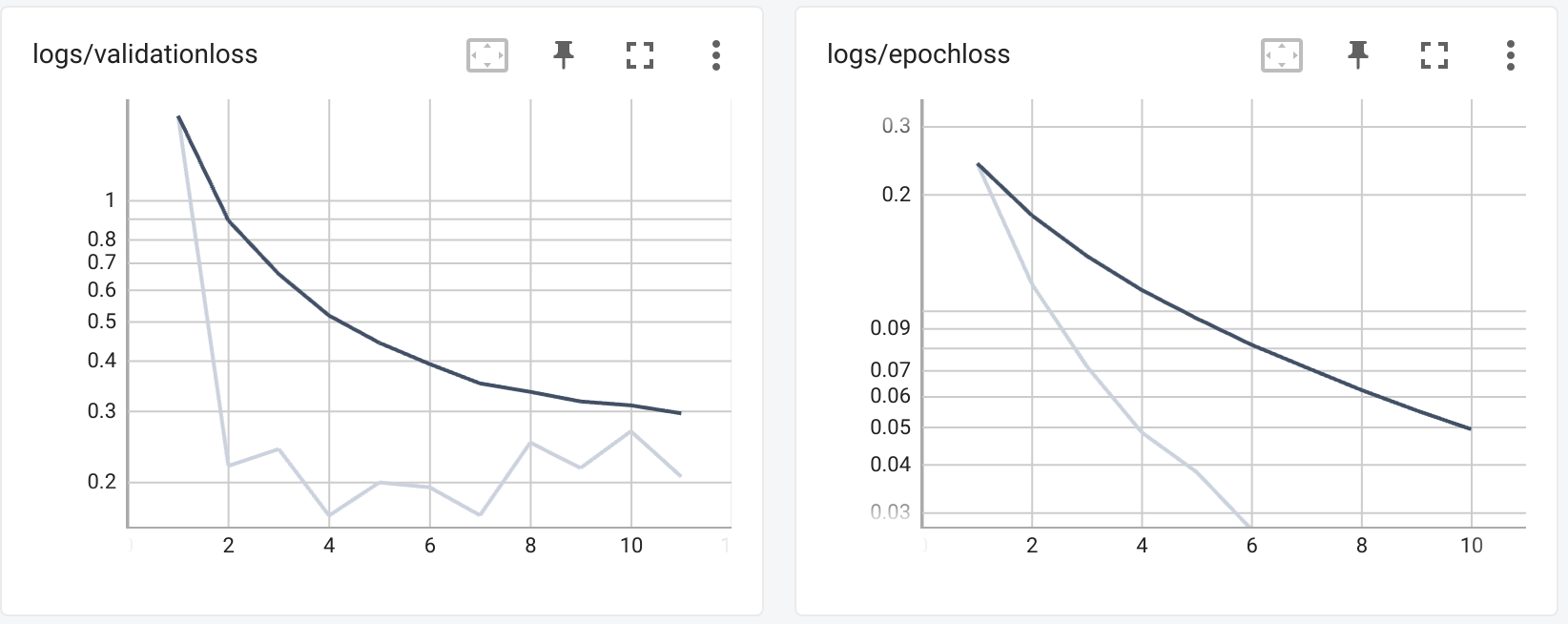}
\caption{Validation and Training Losses}
\label{fig:losses}
\end{figure}

\begin{figure}[H]
\includegraphics[scale=0.15]{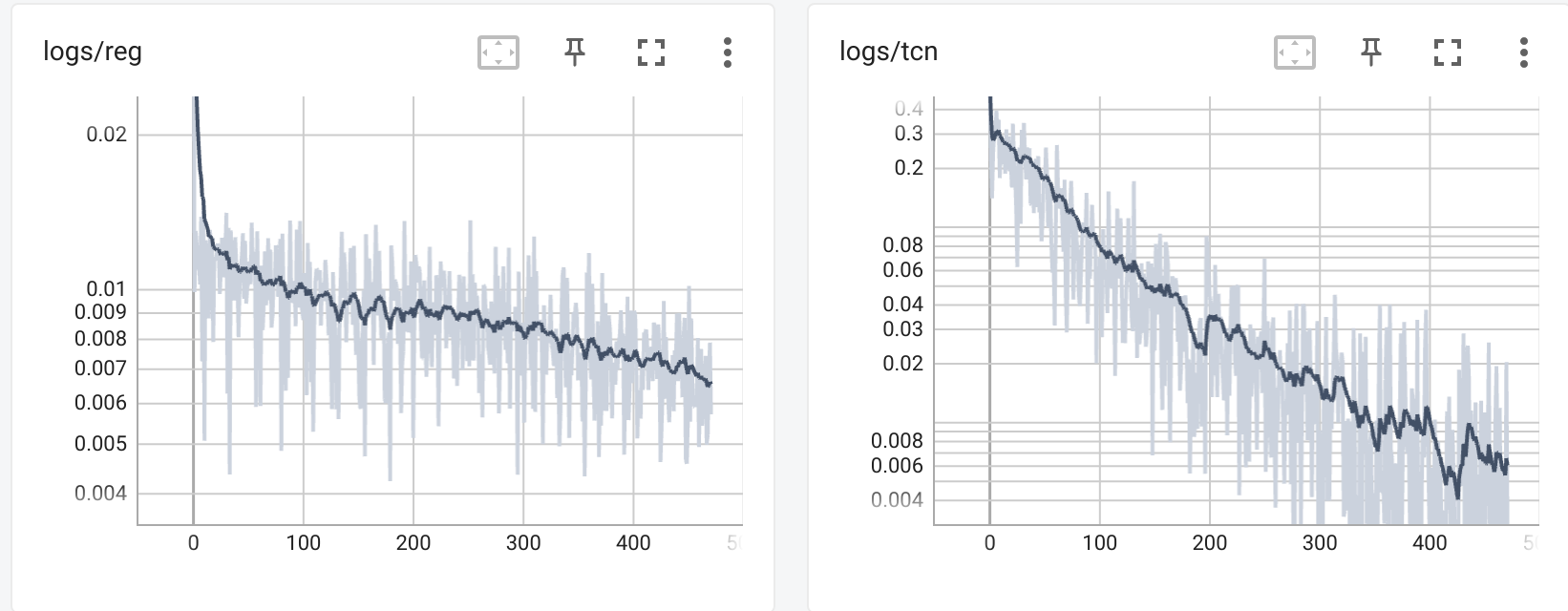}
\caption{Individual Regression and TCN losses during training}
\label{fig:regtcn}
\end{figure}

\begin{figure}[H]
\includegraphics[scale=0.12]{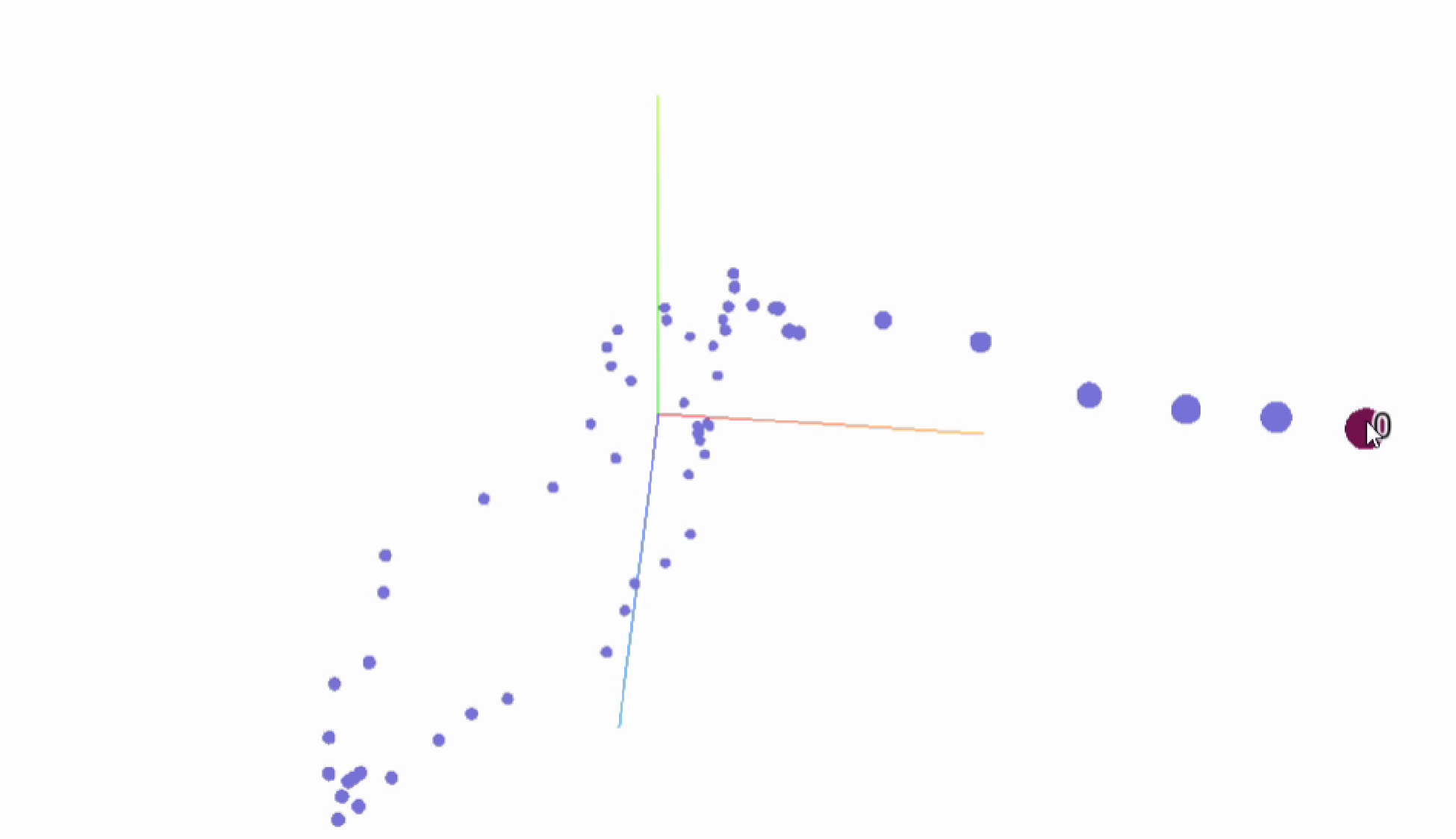}
\caption{TCN Representation of a robot video in 3D PCA}
\label{fig:regtcn}
\end{figure}

\begin{figure}[H]
\hspace{2mm}%   
\includegraphics[scale=0.7]{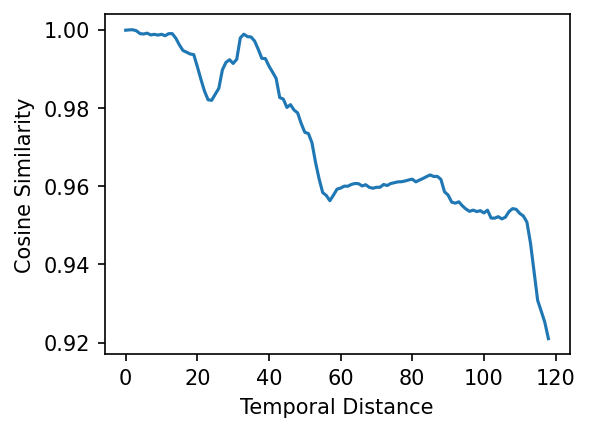}%
\caption{Cosine Similarity of a video}
\label{fig:cosine}
\end{figure}
As I have mentioned in the previous section \ref{workingtcn}, I have created a validation dataset and setup \enquote{early stopping} to prevent overfitting. The network soon converged and began overfitting, due to the small dataset size.
The Fig. \ref{fig:losses} shows the validation loss (in the left) and training loss. It should be noted that the validation loss has plateaued while the training loss kept converging, indicating over-fitting on the training data. I tried to train on a large dataset, but was bottle-necked with limited memory. Recently, I was able to breakdown the dataset into fragments and was able to dynamically load data during the training process. However, It was too late for training as I did not have enough time.

Overall, the error is high (because of the overfitting), and the imitation is not satisfactory.

\section{How Can I Make It Better?}
I strongly believe that increasing the dataset size can significantly reduce the testing error. On top of that, I used domain randomization for every 60 frame. Instead, if I had used domain randomization at every frame, it would have reduced the over-fitting. Due to computational limits, I had to use the worse performing variant of TCN, i.e. Single-view Triplet Loss Implementation, instead of the multiview n---pairs implementation. I believe using the multiview n--pairs implementation can significantly improve the performance, as demonstrated in the original TCN paper.
%\bibliographystyle{ieeetr}
%\bibliography{refs}

\vspace{12pt}
\end{document}